\documentclass[
superscriptaddress,
amsmath,amssymb,
aps,
]{revtex4-2}

\usepackage{soul}
\usepackage{graphicx}
\usepackage{subcaption}
\usepackage[dvipsnames]{xcolor}
\usepackage{amssymb}
\usepackage{amsmath}
\usepackage{mathtools}
\usepackage{amsbsy}
\usepackage{color}
\usepackage{float}
\usepackage{bm}
\usepackage{tikz-cd}
\usepackage{algorithm}
\usepackage{algpseudocode}
\usepackage{xparse,xfp}
\usepackage{hyperref}      

\makeatletter 
\NewDocumentCommand{\acr}{m}{{%
  \fontsize{\fpeval{round(0.8*\f@size,0)}}{\f@baselineskip}\selectfont
  #1%
}}
\makeatother

\usepackage{bm}

\usepackage{siunitx}

\begin{document}

\title{Shortcut learning in geometric knot classification}

\author{Djordje Mihajlovic}
\affiliation{School of Mathematics, University of Edinburgh, Edinburgh, EH9 3FD, UK}
\affiliation{School of Physics and Astronomy, University of Edinburgh, Edinburgh, EH9 3FD, UK}

\author{Davide Michieletto}
\affiliation{School of Physics and Astronomy, University of Edinburgh, Edinburgh, EH9 3FD, UK}
\affiliation{MRC Human Genetics Unit, Institute of Genetics and Cancer, University of Edinburgh, Edinburgh, EH4 2XU, UK}
\affiliation{International Institute for Sustainability with Knotted Chiral Meta Matter (WPI-SKCM$^2$), Hiroshima University, Higashi-Hiroshima, Hiroshima 739-8526, Japan}


\newcommand{\dmi}[1]{\textcolor{RoyalBlue}{#1}}

\begin{abstract}
Classifying the topology of closed curves is a central problem in low dimensional topology with applications beyond mathematics spanning protein folding, polymer physics and even magnetohydrodynamics. The central problem is how to determine whether two embeddings of a closed arc are equivalent under ambient isotopy. Given the striking ability of neural networks to solve complex classification tasks, it is therefore natural to ask if the knot classification problem can be tackled using Machine Learning (ML). In this paper, we investigate generic shortcut methods employed by ML to solve the knot classification challenge and specifically discover hidden non-topological features in training data generated through Molecular Dynamics simulations of polygonal knots that are used by ML to arrive to positive classifications results.  
We then provide a rigorous foundation for future attempts to tackle the knot classification challenge using ML by developing a publicly-available (i) dataset, that aims to remove the potential of non-topological feature classification and (ii) code, that can generate knot embeddings that faithfully explore chosen geometric state space with fixed knot topology. We expect that our work will accelerate the development of ML models that can solve complex geometric knot classification challenges. 
\end{abstract}

\maketitle

\section{Introduction} 

Knot theory, a topic within low-dimensional topology, concerns the study of embeddings of the form $S^{1}\hookrightarrow \mathbb{R}^{3}$,  \emph{knots}. Knots are said to be equivalent if one embedding can be transformed into another under ambient isotopy, i.e. a smooth deformation of the embedding without breaking the curve or passing through oneself. Ambient isotopies can be realized diagrammatically through Reidemeister moves, such that two embeddings are said to be \textit{topologically equivalent} if they can be transformed into each other through a series of Reidemeister moves~\cite{adams2004knot}.

\begin{figure}[t!]
      \includegraphics[width=1\linewidth]{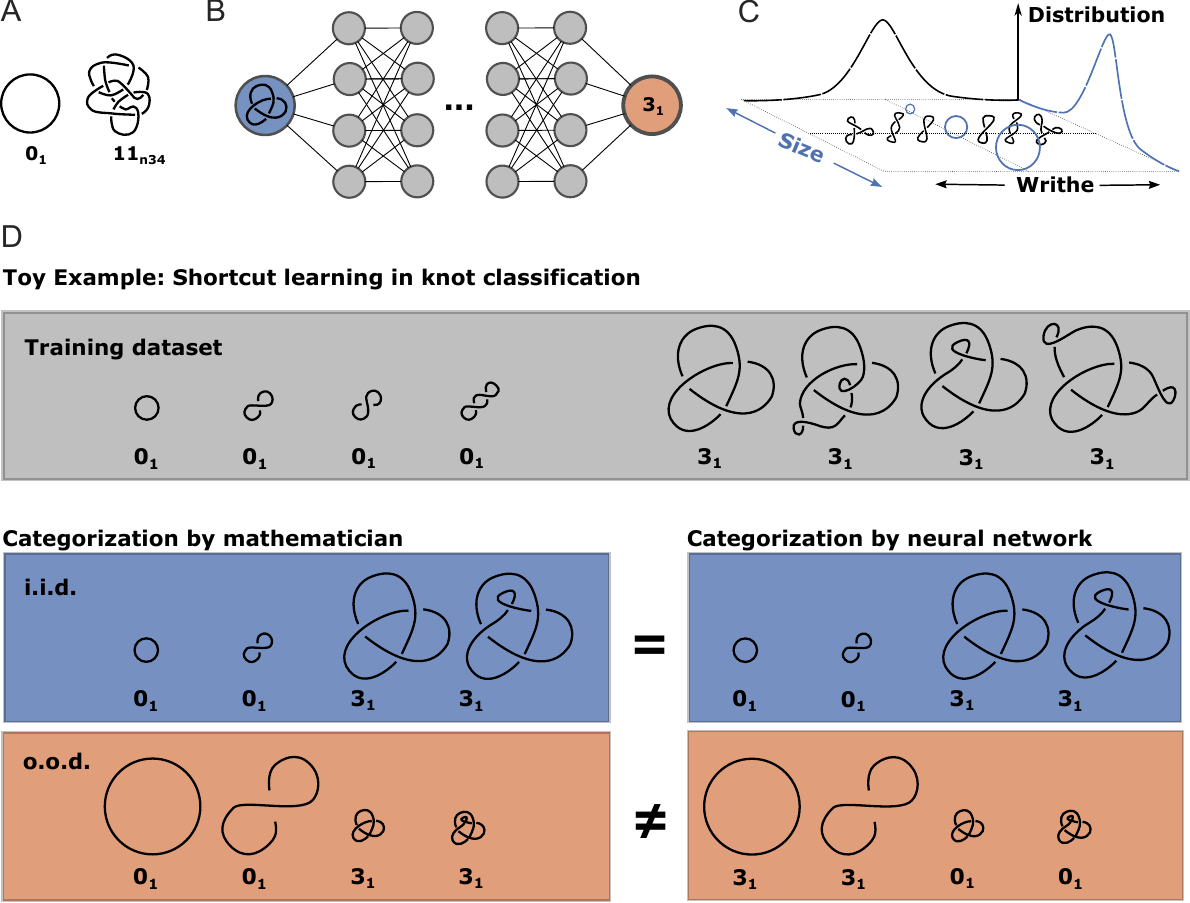}
        \caption{\textbf{Shortcut learning of knot topology.} A.) The unknot ($0_1$) and the Conway knot ($11_{34}$) have the same (trivial) Alexander polynomial. However one cannot be deformed into the other. B.) A sketch of a neural network taking a knot embedding coordinates as input and outputting a binary unknot versus trefoil ($0_1$ vs.\ $3_1$) classification. C.) An example of a configurational landscape of a knot embedding. The two main directions here are ``size'' and ``writhe'', i.e. the amount of self-crossing of the curve. Molecular Dynamics simulations intrinsically bias sampling towards low free energy embeddings and therefore generate knot conformations with narrow distributions of geometric properties. D.) Example of a training dataset in which the ML takes an obvious shortcut learning based on the size of embeddings. All the trivial knots ($0_1$) are smaller than the trefoils ($3_1$) in the dataset. When the ML is challenged with $0_1$ that are large, and $3_1$ that are small it leads to incorrect classification. Figure layout inspired from~\cite{geirhos2020shortcut}.}
    \label{fig:introduction}
\end{figure}

The challenge of classifying knots up to topological equivalence dates back to Peter Guthrie Tait who was inspired by Lord Kelvin's idea of knotted vortex atoms~\cite{przytycki1998classical}. In current tabulations, different knots are labelled as $P_{Q}$, where $P$ denotes the minimal number of crossings in any projection of a knot, and $Q$ denotes a (somewhat arbitrary) order within knots with same crossing number. Beyond 10 crossings, an additional labelling is conventionally added to denote properties such as alternating, $a$, or non-alternating, $n$, giving $P_{nQ}$ (Fig. \ref{fig:introduction}A).

Knots are classified by topological invariants, i.e. quantities that can be computed on 3D embeddings or their 2D projections, and that are invariant under ambient isotopy, such as Reidemeister moves. However, finding a ``complete invariant'', i.e. one that can uniquely classify any two knots, remains an open problem in mathematics~\cite{adams2004knot}. Indeed, mathematicians have developed several invariants spanning both geometric and topological construction, for example the Jones polynomial~\cite{jones1987hecke}, HOMFLY-PT polynomial~\cite{freyd1990new}, hyperbolic volume~\cite{thurston2022geometry}, and Vassiliev invariants~\cite{vassiliev1990cohomology}. However, each invariant developed thus far is not known to be able to uniquely identify every knot, i.e. it is so far always possible to find two topologically distinct knots that share the same topological invariant. An enlightening example is the comparison of the unknot ($0_1$) with the Conway knot ($11_{n34}$), drawn in Figure ~\ref{fig:introduction}: the unknot cannot be smoothly deformed into the 11 crossing Conway knot, however, they both have the same (trivial) Alexander polynomial.

In the last $\sim$10 years, machine learning (ML) has become an ideal tool to learn complex patterns and solve classification tasks. To do so, an ML model learns a function $f$ from labeled data $X^{(j)}=\{ x_{1}, ..., x_{i} \}, y^{(j)}$, such that $f: X^{(j)} \rightarrow y^{(j)}$. ML techniques therefore lend themselves to be applied to problems in knot theory and have shown some preliminary success in both classification and conjecture generation~\cite{gukov2021learning,Davies2021,jejjala2019deep,craven2023learning, gukov2023searching}.

In addition to the mathematical challenge of classifying knots based on their embedding, knots in biophysical systems such as DNA, proteins, polymers, or fields display additional physical constraints that are manifested through the specific geometric embedding of the knotted object~\cite{Ricca2024KnottedFields,stasiak1998ideal}. Importantly, some geometric features such as protein folds, are important for the function of such biophysical molecule~\cite{Tubiana2024Topology}. It is thus intriguing to conjecture that physical embeddings have properties which are reflected by the underlying topology of the object. The hope is that we can discover quantities that connect functionally important geometric features to structurally informative topological motifs in knotted biological and physical matter~\cite{Tubiana2024Topology}.  

Motivated by this idea, several papers have proposed ML models to analyse embeddings of polygonal knots using geometric features, such as coordinates of the discretized segments and other geometrically-inferred measurements such as curvature and local writhe~\cite{vandans2020identifying, braghetto2023machine, sleiman2024geometric, zhang2025recognizing, braghetto2025variational}. Despite the fact that these features are not topologically invariant, supervised ML models appeared to successfully solve knot classification tasks (with $> 99\%$ accuracy), in turn sparking the question of whether ML models are ``learning'' topological invariants from algebraic patterns within these non-topological, geometric features. 

The idea underlying these attempts is that there exists many known connections between the topology of a curve and geometric properties of its embedding. For instance, the Fary-Milnor theorem states that if the total curvature of an embedding is less than $4 \pi$, then the curve must be unknotted~\cite{Tubiana2024Topology}, thus using geometry to provide a (weak) bound on the topology. Another example is the White-Fuller-C\u{a}lug\u{a}reanu's theorem,
\begin{equation}\label{eq:linking}
    \text{Lk}(K) = \text{Wr}(K) + \text{Tw}(K)
\end{equation}
stating that the sum of writhe (\text{Wr}) and the twist (\text{Tw}) of a (framed) knot -- both affected by ambient isotopy -- yield a formal topological invariant that is independent of smooth deformations and Reidemeister moves. Finally, a more biological example where topology is intimately connected to geometry is the case of DNA knots moving through a porous structure during gel electrophoresis~\cite{Katritch1996a}. The speed at which knotted DNA travels through the gel depends on the volume occupied by the molecule in space, itself dependent on the topology of the DNA~\cite{Katritch1996a,Arsuaga2002,Trigueros2007,Michieletto2015Topological,Valdes2019QuantitativeDNAknot}. Similarly, embedding-dependent geometric measures, like space writhe, have been shown to correlate with topological invariants in physically realistic ensembles~\cite{thompson2025space}.

Motivated by these results, here we aim to evaluate the success of ML models used to classify polygonal knots and analyse their performance in light of \emph{shortcut learning}. Shortcut learning refers to a process in which the supervised ML model picks up unwanted features of the training dataset and correlates them with the label in order to solve the classification task.  Extreme examples of this process are NNs learning to classify images of cows based on green hills in the background~\cite{beery2018recognition} or misclassifying images of animals based on their texture~\cite{geirhos2018imagenet}. However, shortcut learning can be more nuanced and, as we show in this paper, can lead to inaccurate interpretations, which in turn provides enlightening steps to understanding classification of topology in biological, energetically constrained, systems. 
Investigating shortcut learning in topological classification will improve our understanding of ML for mathematics and physics, and increase the likelihood of building future models capable of learning true topological information based on geometric information.

To better understand whether shortcut learning affects existing ML models, in this paper, we consider NNs trained to solve a reduced knot classification task, detecting trivial knots ($0_1$) versus a chosen non-trivial knot ($3_1$, Fig.~\ref{fig:introduction}B) and we develop a process to infer potential shortcuts by computing mutual information scores based on a variety of geometric measures in the input data set. 
We argue that this reduced setting is sufficient for our purposes. Specifically shortcut learning, if present, must already be present at the level of distinguishing the simplest non-trivial knot from the unknot. Any non-topological, geometric feature exploitable in this setting could persist, and potentially compound, in higher crossing regimes.
Our method to flag potential shortcuts does not directly inform us on the shortcut learning process; however, it flags features of the input data as potentially being learned. For instance, we show that knots generated by Molecular Dynamics (MD) simulations (as in Refs.~\cite{vandans2020identifying, braghetto2023machine, sleiman2024geometric, zhang2025recognizing, braghetto2025variational}) return high mutual information scores, likely caused by limitations in bending stiffness, length and the energetics of the simulation, which greatly constrain the space of geometric conformations explored. These constraints result in narrow conformational distributions of the input datasets and therefore potential for topological shortcut learning based on geometric landscapes (Fig.~\ref{fig:introduction}C). To solve this issue, in this paper, we also introduce \acr{GEOKNOT}, an open-source package to sample polygonal knots in user-defined regions of the geometric landscape. We demonstrate that models with high knot classification accuracy on MD generated knots do not perform as well on \acr{GEOKNOT} generated data, further supporting the existence of shortcut learning in these models.

This paper is structured as follows: in section ~\ref{sec:methods} we define the method and set of observables that can be computed on input data to diagnose the likelihood of them leading to ``shortcut learning'' of the ML model and report details of the ML models considered in this work. 
In Sec.~\ref{sec:sim_methods} we will present the algorithms employed to generate the input datasets, in both the MD sense as prescribed in~\cite{vandans2020identifying, braghetto2023machine, sleiman2024geometric, zhang2025recognizing, braghetto2025variational} and via our new knot sampling tool \acr{GEOKNOT}.
In Sec.~\ref{sec:results} we will present the results of our experiments; briefly, we discover that knots sampled with geometrically unbiased algorithms cannot be accurately classified using ML models that previously demonstrated high classification accuracy on the geometrically constrained samples. We argue that the original MD trained models undertake a shortcut learning based on geometric features of the knots which do not correlate with the underlying topology in o.o.d. (out of distribution) settings. Finally in Sec.~\ref{sec:discussion} we summarise our findings and suggest future directions for improving the community's efforts to train ML models to ``learn'' topological invariants in a geometric setting.

We stress that this work does not claim that ML models are incapable of learning knot topology; rather, we investigate the importance of sampling data and find that common methods employed in polygonal knot classification intrinsically correlate geometric features that are not inherently topologically invariant. Our aim is therefore not to discount prior results, but to clarify the distinction between situationally powerful geometric probes and genuine topological classification, and to shed light on ML interpretability in this domain.

\section{Methods}
\label{sec:methods}

In this work we consider each datapoint, $x$, to represent a single polygonal knot embedding (e.g. a discretized, closed, non-intersecting curve in $\mathbb{R}^3$). Each datapoint has attached a label, $y$ which encodes the topology, or ``knot class'' (i.e., either $0_1$ or $3_1$ in this work). Our ML models will then be trained on geometric input, the classification task can essentially be solved in one of two ways: (i) the model learns an algebraic manipulation of the data to find an ambient isotopic invariant measure or a ``\emph{topologically faithful}'' decision rule or, alternatively, (ii) the model learns to solve the classification task based on features of the datasets that are not sufficiently expressed across all knot classes. For example, the ML model could easily learn to classify knots if the size occupied in 3D space of the polynomial curves was systematically larger for the unknotted curves than for the knotted ones (see Fig.~\ref{fig:introduction}D). In this case, while the model would easily achieve a very high classification accuracy, it would not have learned anything about the topology of the curves, because the 3D size of the knots is not invariant under ambient isotopy. Instead, it would have learned a simple correlation between the label $y$ and the geometric size of the input curve. 

Learning to solve a classification task by using features that are correlated with the labels but not conceptually meaningful is referred to as \emph{shortcut} learning. In the specific case of knot classification, we call a \emph{shortcut} any geometric functional $\phi(x)$ that is not intended to be a measure of topology and is therefore not invariant under ambient isotopy. However, such geometric functional $\phi(x)$ could still be statistically correlated to the knot type and might therefore be used by the model to shortcut learn. 

A model trained on such data would likely achieve high in-distribution classification accuracy, creating the false assumption that the ML network has `learned topology' in the strict sense. Importantly, shortcut dependence is a property of the interaction between a model and the corresponding dataset; the same architecture may or may not rely on shortcuts depending on their availability in the dataset. 

Testing shortcut learning is in general difficult due to the ``black box'' nature of ML models. We therefore assess ML models that have displayed high accuracy in knot classification task by introducing a diagnostic tool at the level of datasets, i.e. a ``shortcut'' probe. Such a probe will be aimed at identifying features that strongly correlate with labels, but are not invariant under ambient isotopy and can therefore lead to shortcut learning. 

The shortcut probe is constructed as follows: given a representation map $r(x)$ (e.g., 3D coordinates of the polygonal segments) and a collection of geometric functionals $\Phi=\{\phi_j\}_{j=1}^k$ (with $k$ the number of different functionals, see next section), we score the mutual information between each geometrical functional $\phi_j$ and the labels $y$ in the dataset. If the mutual information is large, then $\phi_j$ is highly correlated with the label. 

This method is commonly used in ML for feature engineering~\cite{vergara2014review}, however we here employed it to detect shortcut learning. We emphasise that this procedure is intentionally conservative: a high mutual information score does not prove that a model will use that feature to learn, and a low score does not rule out additional complex shortcuts; rather, this procedure identifies geometric observables that are \emph{sufficient} within the given dataset to support positive classification.


\subsection*{A diagnostic probe for  shortcut learning in topology classification}

To diagnose geometric shortcut learning in topology classification tasks, we develop quantities that detect purely geometric features that may correlate with knot topology. We propose the following ``shortcut probe'' algorithm:
\begin{algorithm}[H]
  \caption{Shortcut Probe}\label{alg:gap-over-N}
  \begin{algorithmic}[1]
    \Require Dataset $D = \{ x^{(i)}, y^{(i)} \}_{i=1}^{n} $, representation $r(x)$ for each $x$, collection of feature functionals $\Phi = \{\phi_{j}\}_{j=1}^{k}$.
    
    \State $\mathcal{R} \gets \varnothing$ 
    \For{$x, y \in D$ }
          \State $z \gets r(x)$  
          \For{$\phi_{j} \in \Phi$}
                \State $v_{j} \gets v_{j}\cup \phi_{j}(z)$
          \EndFor
    \EndFor
    \For{$j\in (1, k)$}
          \State $s_{j} \gets I(v_{j}, y)$
          \State $\mathcal{R} \gets \mathcal{R} \cup \{ (\phi_{j}, s_{j}) \}$
    \EndFor

    \Return $\mathcal{R}$
  \end{algorithmic}
\end{algorithm}

Algorithm \ref{alg:gap-over-N} returns probes of given datapoints within the dataset; then, mutual information is computed as
\begin{equation}
\label{eq:mutinf}
    I(X; Y) = \sum_{y\in Y}\sum_{x\in X}p(x, y)\text{log}\frac{p(x, y)}{p(x)p(y)} \, ,
\end{equation}
and it provides a metric for determining, given $X$, the uncertainty in label $Y$. In other words, $I(v_j; y)$ in the shortcut probe algorithm quantifies the likelihood that different geometric quantities (computed through the functionals $\phi_j$) may be used to solve the classification challenge. 

In practice, to compute mutual information with continuous $X$ and discrete $Y$, we implement readily available methods from Scikit-learn~\cite{pedregosa2011scikit} developed using non-parametric methods based on entropy estimation from k-nearest neighbour distances~\cite{ross2014mutual, kraskov2004estimating}. 
To utilize this probe in a knot classification challenge, we choose various functionals ($\Phi = \{\phi_{j}\}$) that capture geometric properties of polygonal curves, e.g. variation of the pairwise distance matrix $M_{ij}$ between segments in the polygonal knot, total sum of space writhe and polygonal curvature. Additionally, scalar values such as sum over pairwise distance matrix ($\Sigma_{+}$) and sum of maxima at different tolerances, $n$, across $M_{ij}$ ($\Pi_{n}$) are also considered. We summarise and define the functionals used in this work in Table \ref{tab:functionals}.
\begin{table}[hbt!]
    \centering
    \begin{tabular}{c|c}
     $\phi_{j}(\bm{x})$ & Defn. (wrt. $\bm{x}$) \\ [0.5ex] 
         \hline\hline
        $\mathbf{\Sigma}_{+}$&  $\sum_{i, j}|\bm{x}_{i}-\bm{x}_{j}|$ \\
        $\mathbf{\Omega}_{+}$& $ \sum_{i,j} (d\bm{x}_i \times d\bm{x}_j) \cdot (\bm{x}_i - \bm{x}_j)/|\bm{x}_i - \bm{x}_j|^3 $ \\
        $\boldsymbol{\kappa}_{+}$& $\sum_{i} \text{arccos}((\bm{x}_{i} \cdot \bm{x}_{i+1}) / (|\bm{x}_{i}| |\bm{x}_{i+1}|))$ \\
        $\textbf{M}$& $\text{Max}(|\bm{x}_{i}-\bm{x}_{j}|)$ \\ $\mathbf{\Pi}_{n}$ & $\sum_{i, j}\pi_{ij}, \pi_{ij} \in M_{ij} | \text{ connected component}>n$
    \end{tabular}
    \caption{Functionals ($\phi_{j}$) used to identify potential shortcut learning. Here, $\bm{x}_{i}=r(x_{i})$ represents the 3D coordinates of vertex $i$ in the polygonal knot. From top to bottom: $\mathbf{\Sigma}_{+}$ is the sum over pairwise distances of segments, $\mathbf{\Omega}_{+}$ is the total writhe, $\boldsymbol{\kappa}_{+}$ is the total curvature, $\textbf{M}$ is the largest pairwise distance and $\mathbf{\Pi}_{n}$ is the number of peaks in the pairwise distance matrix, such that the peaks are higher than the tolerance $n$. }
    \label{tab:functionals}
\end{table}

From the calculation of the shortcut probe, it is natural to define a ``shortcut index'', $\tau$, computed by comparing ML models trained on the original features with those trained solely on the shortcut features flagged by our probe. 
The shortcut index quantifies the likelihood that a model has learned to solve a knot classification challenge using these specific shortcut probes. To practically compute $\tau$, we train models using features scoring highly on the shortcut probe as input, as opposed to coordinate data and/or some other input. The accuracy of these models, $m_{a}$ is compared to the accuracy of models trained on the untransformed data, $m$, and $\tau$ is subsequently defined as the ratio $m_{a}/ m$. Importantly, $\tau$ is computed using fixed architecture and training setup such that $m$ and $m_{a}$ differ only in the feature space of the input data used. This definition of $\tau$ allows us to quantify the extent to which geometric features suffice to explain a large fraction of the model performance, serving as an indication of shortcut learning.

\subsection*{Details of ML model architecture, training and testing } 
We train models using simple feed-forward neural networks in PyTorch~\cite{paszke2019pytorch}.  Unless otherwise specified, the models consist of 4 layers with 320 neurons through which values are passed after the required flattening; for example, for coordinate based data, flattening is per coordinate such that structure follows $x_{1}, y_{1}, z_{1}, x_{2}, ..., z_{n}$, matrices follow similar logic, flattening by row. Additionally, we centre each configuration by subtracting its centroid and normalize scale to avoid spurious cues from absolute position and global scale. Adam optimisation~\cite{kinga2015method} is used to ensure optimal training with a $10^{-3}$ learning rate. Categorical cross entropy loss allows accurate evaluation of the models during training. Finally, model weights are initialised using standard He-Initialisation ~\cite{he2015delving}. We choose consistent and simple models to replicate the results reported in Ref.~\cite{sleiman2024geometric}, where it was found that this architecture can faithfully classify up to 250 different knot types with 95\% accuracy. Arguably, if simple FFNN models can achieve high accuracy using only shortcut features, then more complex architectures trained on similarly generated datasets (as in Refs.~\cite{vandans2020identifying, braghetto2023machine, braghetto2025variational, sleiman2024geometric,zhang2025recognizing}) should have the capacity to exploit the same shortcuts. Thus, the shortcut index $\tau$ provides a conservative lower bound on shortcut reliance for more complex models trained on the given data.

\subsection*{Dataset creation and main input representations} \label{sec:sim_methods}
For this experiment, we consider a simplified classification challenge with respect to previous works. Specifically, we restrict our datasets to an ensemble of unknots ($0_1$) and trefoil ($3_1$) knots. We argue that this simplified challenge is sufficient to demonstrate the potential for shortcut learning in generic knot classification tasks.
We create datasets containing at least 1000 different embeddings of either knot class. Both datasets are shuffled using a chosen seed for reproducibility and a [train : val : test] split of [0.8 : 0.05 : 0.15] is used in all training instances. We train models broadly on two input representations: (i) 3D coordinate data of each segment in the polygonal knot and (ii) ``writhe matrix'', $\Omega_{\alpha\beta}$, which represents the discretized space writhe density for each polygonal knot computed as~\cite{banchoff1976self}:
\begin{equation}\label{eq:ab_prop}
   \Omega_{\alpha\beta} = \dot{r}^{i}(x_{\alpha})\dot{r}^{k}(x_{\beta})\epsilon_{ikm}\frac{(r(x_{\alpha}) - r(x_{\beta}))^{m}}{4\pi|r(x_{\alpha}) - r(x_{\beta})|^{3}}
\end{equation}

In this work we also compare three main datesets, two created using Molecular Dynamics simulations in LAMMPS, and one created using a Monte Carlo algorithm. 

\subsubsection*{Molecular Dynamics sampling of polygonal knots}\label{sec:LAMMPS}

Currently, the majority of published work on ML classification of knots have adopted a common data generating strategy that employs Molecular Dynamics simulations. Here we consider a dataset from Ref.~\cite{sleiman2024geometric} which is generated by initializing desired polygonal knots using KnotPlot~\cite{scharein2002interactive} and representing them as closed bead–spring chains of length $N$. The position of each bead (vertex) is then updated through a Langevin equation subject to potentials accounting for excluded volume between pairs of beads, bonds between consecutive beads along the chain and finally ``angle'' potentials between consecutive triplets of beads~\cite{sleiman2024geometric}. The parameters of these potentials and the integration timestep ($dt = 0.01\tau_{\mathrm{LJ}}$) are set in such a way to preserve the chain topology at all times~\cite{Kremer1990} (and we check that the topology is preserved by tracking Alexander determinant during the evolution of these chains). The evolution of the Langevin equation is performed at fixed number, volume and temperature and using a velocity–Verlet scheme in LAMMPS~\cite{Plimpton1995a,sleiman2024geometric}. By saving the configuration of the chains every $10^7 dt = 10^5 \tau_{\mathrm{LJ}}$ it is possible to obtain statistically and thermally uncorrelated conformations of topologically identical curves.

Existing works differ mainly in physical regime rather than modelling principle. In Refs.~\cite{braghetto2023machine,braghetto2025variational} fully flexible rings are sampled under strong spherical confinement, producing dense globular conformations, whereas Refs.~\cite{sleiman2024geometric,zhang2025recognizing} generate configurations without external confinement, including semi-flexible and flexible polymers at fixed chain length and temperature. In all cases, however, datasets arise from bead-spring polymer models with fixed parameters (chain length, stiffness, confinement, density), subject to energy constraining potentials. Within these simulations, the exploration of the space of geometrical embeddings of a knot is constrained by energy potentials, in turn yielding relatively narrow distributions of, for instance, geometric size $R_g^2 = (1/N^2) \sum^N_{ij} (\bm{x}_i - \bm{x}_j)^2$ and overall writhe $Wr = \oint_1 \oint_2 (d\bm{x}_i \times d\bm{x}_j) \cdot (\bm{x}_i - \bm{x}_j)/|\bm{x}_i - \bm{x}_j|^3 $ of the knot embeddings. In other words, since both of these features are controlled by chain flexibility and length, performing temperature-conserving simulations restricts the space of $R_g$ and $Wr$ that can be explored by using NVT MD simulations.

As a mean of comparison, in our work we sample knots using LAMMPS at two different temperatures $T=1$ (low) and $T=10$ (high) to speed up exploration of embedding geometries. Specifically the low temperature, $T=1$, was employed in all of the current literature~\cite{braghetto2023machine,braghetto2025variational, sleiman2024geometric,zhang2025recognizing} whereas high temperature, $T=10$, has been chosen to broaden the range of sampled conformations while preserving knot topology. 

\subsubsection*{Biased Monte Carlo sampling of polygonal knots: \acr{GEOKNOT}} 

To create a dataset with minimized potential for shortcut learning, we developed our own Markov Chain Monte Carlo sampler called `\acr{GEOKNOT}'. This sampler allows the user to bias the range of sampling in state space and therefore to efficiently sample random embeddings with chosen geometric features. We note that the evolution of conformations is done on a lattice, because both evolution and computation of geometric properties of lattice knots is much faster than that of off-lattice knots~\cite{cimasoni2001computing}. Specifically, we employ the BFACF algorithm~\cite{de1983new} to perform local topology-preserving moves, ensuring ergodic sampling within each knot type~\cite{van1992ergodicity}. We complement this with pivot moves~\cite{madras1988pivot}, which introduce efficient non-local geometric rearrangements. Pivot moves allow for faster geometric exploration at the expense of topological invariance. To maintain topological consistency, the topology of each configuration is verified at each move using the Alexander polynomial, computed using KymoKnot~\cite{tubiana2018kymoknot}. We then introduce a small displacement controlled by a constant random seed to move the knot off lattice and produce the final embedding. Since the Alexander polynomial is not a unique knot invariant, we also perform secondary checks of the second and third Vassiliev invariants to further confirm the knot class of the curve at each Monte Carlo iteration. 

We assume that this combination of invariants is powerful enough to ensure consistent topology, as examples which share all three invariants are incredibly rare and do not occur within the first 10 crossing ($> 250$) prime knots. However, to minimize any possible error we randomly test a small subset of the topology of samples in \acr{GEOKNOT} by energetically minimizing the configuration in KnotPlot~\cite{scharein2002interactive} - this simplifies the 3D configuration such that visual inspection is enough to determine topology.

On top of these moves, we explicitly bias configurations to explore a broad range of geometric features. For example, we force total space writhe, average crossing number, and long-range entanglement to explore a broad, flattened distribution in state space. We do this by tracking all possible geometric features during the proposed move, and by applying a sampling bias that favours empty regions, accepting only proposed configurations that fill a user-defined region of a given geometric property. Effectively, our algorithm constantly updates the bias, acting as a variant of the traditional Wang-Landau sampler~\cite{landau2004new}. 

As a result, \acr{GEOKNOT} generates conformations that are uncorrelated and display controllably broad distributions along the desired geometric directions. Each configuration in \acr{GEOKNOT} is validated for self-avoidance and topology preservation and exported as 3D coordinate sequences suitable for direct use in ML pipelines. Full algorithmic details and metric definitions are provided in Appendix~\ref{appendix:A} and Appendix~\ref{appendix:B} respectively. A limitation of \acr{GEOKNOT} is that the sampling time becomes longer if the user desires more complex knot classes. For example, a 11-crossing knot will be sampled less efficiently than a 3 crossing knots because of the larger chances of rejection of topology-altering pivot moves. In addition to this, computation of polynomial invariants is expensive, acting as a bottleneck at scale; recent work considers parallel algorithms for computation of Jones polynomials which could be beneficial in this domain~\cite{barkataki2025parallel}. Consequently, the analysis of this data considers sampling between only unknots ($0_1$) versus trefoil knots ($3_1$). By saving the configurations once they fill an empty bin requirement, we generate a dataset with at least 1000 uncorrelated conformations. Because not constrained by energy potentials and temperature, and attempting to broadly sample in geometric space, this dataset is expected to lead to a smaller likelihood of shortcut learning.
\section{Results} 
\label{sec:results}

\subsection*{ML models fail to classify the \acr{GEOKNOT} dataset}
\begin{figure}[ht!]
\includegraphics[width=0.9\linewidth]{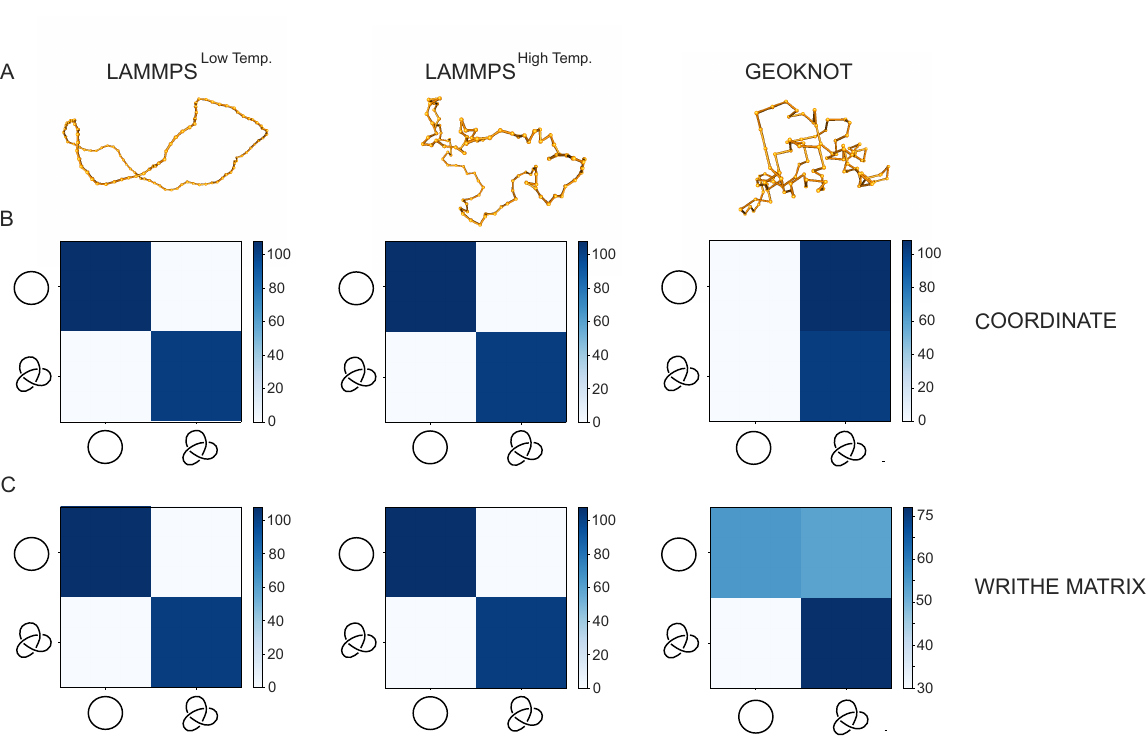}
    \caption{\textbf{ML models fail to classify \acr{GEOKNOT} dataset.} \textbf{A} Examples of knot embeddings from MD simulations at low and high temperatures and of \acr{GEOKNOT}. All three embeddings have 100 nodes. Plotted using KnotPlot~\cite{scharein2002interactive} \textbf{B} Confusion matrices for models trained on $\text{MD}^{\text{Low T}}$, $\text{MD}^{\text{High T}}$, and \acr{GEOKNOT} respectively with coordinate data as the feature input. \textbf{C} Confusion matrices for models trained on $\text{MD}^{\text{Low T}}$, $\text{MD}^{\text{High T}}$, and \acr{GEOKNOT} respectively with writhe matrix data as the feature input.}
    \label{fig:figure_2}
\end{figure}

First, we generate three datasets as described in Sec.~\ref{sec:methods}: two via MD with high and low temperatures and one using \acr{GEOKNOT}. All three datasets are made of 1000 instances per knot class and each embedding has constant polygonal length of 100 vertices (see Figure~\ref{fig:figure_2}A for examples).
We then train and test ML models to classify knots in these datasets using two types of data representation: 3D coordinates of the embedding segments (as in Refs.~\cite{vandans2020identifying, zhang2025recognizing, braghetto2025variational}) and the writhe matrix representation (as in Ref.~\cite{sleiman2024geometric}). 

Models trained on LAMMPS datasets and tested on LAMMPS datasets return the previously published high accuracies, with more than 99\% of conformations classified correctly in both high and low temperatures datasets. However, models trained on the \acr{GEOKNOT} dataset do not perform as well (see confusion matrices in Fig.~\ref{fig:figure_2}B-C), irrespectively of the input representation. 

Finally, models trained on MD data and subsequently tested on \acr{GEOKNOT} embeddings also returned low classification accuracy ($\simeq 50\%$ on coordinate trained models and $\simeq 70\%$ on writhe matrix trained models). 

These results suggest that \acr{GEOKNOT} dataset is more difficult to classify, and that the models trained via MD datasets struggle to generalise to other types of knot embeddings. 
One obvious cause for the poor learnability of the \acr{GEOKNOT} dataset may be the greater geometric complexity (e.g. see snapshots in \ref{fig:figure_2}A). However, as we shall discuss in the next sections, the main reason resides in the absence of geometric features that strongly correlate with knot topology in the  \acr{GEOKNOT} embeddings.  


\subsection*{Shortcut probes detect correlation between geometric embedding and knot topology in MD sampled knots} 

To better understand why ML models that return high accuracy with MD simulated knots fail to classify \acr{GEOKNOT} data, we analyse (i) the distributions of geometric quantities such as writhe across the different datasets (see Appendix~\ref{appendix:B} for details) and (ii) the mutual information between the geometric functionals $\phi_{j}$ (see Algorithm~\ref{alg:gap-over-N}). 

\begin{figure}[ht!]
    \includegraphics[width=0.95\linewidth]{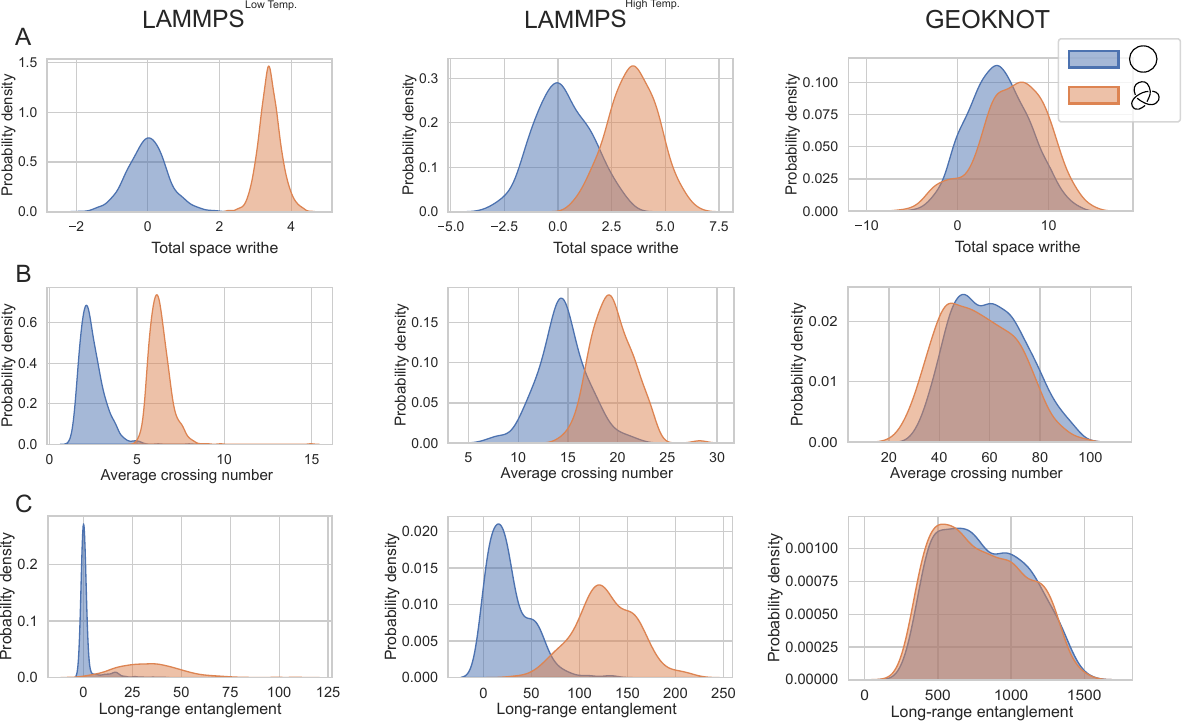}
    \caption{\textbf{MD datasets display limited sampling of geometries.} Distributions of \textbf{A} total writhe, \textbf{B} average crossing number and \textbf{C} long-range entanglement for unknots ($0_1$) and trefoil ($3_1$) knots. From left to right we show distributions from MD low temp, high temp and \acr{GEOKNOT} datasets. The definitions of these quantities are in Appendix~\ref{appendix:B}.}
    \label{fig:figure_3}
\end{figure}

First, we observe that MD generated datasets display narrow distributions of total space writhe, average crossing number, and long-range entanglement (see Figure~\ref{fig:figure_3}, left column). The distributions broaden in the high temperature MD dataset, however they remain clearly separable for the two knot classes (see Figure~\ref{fig:figure_3}, middle column). Indeed, despite the high temperature MD-generated samples display broader writhe values, this is often achieved through small, short-range perturbations along the embedding, e.g. short twists as opposed to long-range coiling.

This observation suggests that MD embeddings span a geometrically constrained portion of the state space. Indeed, despite polygonal knot sample conformations being statistically independent -- as captured at times longer than the relaxation time of the embedding -- they still cover a relatively small region of all possible conformations at fixed topology.


In contrast to the MD generated datasets, the \acr{GEOKNOT} dataset displays very broad distributions of geometric conformations. Importantly, the distributions of these key geometric quantities display significant overlap between the knot classes (see Figure~\ref{fig:figure_3}, right-most column).

To assess the effect of the geometric constraint on the learning process, we then compute the mutual information $I(X;Y)$ (see eq.~\eqref{eq:mutinf}) with $X$ one of the previously defined functionals ($\Sigma_+, \Omega_+, \kappa_+, \cdots$) and $Y$ being the knot type. Through this calculation, we more precisely quantify the correlation between each geometric functional and the knot type. The result of this procedure is summarised in Table~\ref{tab:artefact}, where it is clear that our shortcut probes flag several closely correlated geometric features within the MD data.

\begin{table}[hbt!]
    \centering
    \begin{tabular}{c|c|c|c|c|c|c|c}
     \textbf{DATA} & $\mathbf{\Sigma}_{+}$ & $\mathbf{\Omega}_{+}$ & $\bm{\kappa}_{+}$ & $\textbf{M}$ &  $\mathbf{\Pi}_{5}$ & $\mathbf{\Pi}_{10}$ & $\mathbf{\Pi}_{20}$\\ [0.5ex] 
         \hline\hline
        $\text{MD}^{\text{Low T}}$ &  0.65 & 0.69 & 0.02 & 0.57 & 0.083 & 0.53 & 0.45 \\
        $\text{MD}^{\text{High T}}$ & 0.45 & 0.47 & 0.00 & 0.28 & 0.25 & 0.02 & 0.05 \\
        \acr{GEOKNOT} & 0.02 & 0.03 & 0.05 & 0.01 & 0.02 & 0.01 & 0.00
    \end{tabular}
    \caption{\textbf{Mutual information reveals correlation between geometric features and knot type.} Values of mutual information $I(X;Y)$ for different geometric features compared between MD low temp, high temp and \acr{GEOKNOT} datasets. From left to right: sum of pairwise distances ($\Sigma_{+}$), sum of total space writhe ($\Omega_{+}$), total curvature ($\kappa_{+}$), maximal pairwise distance ($\text{M}$), and the number of peaks on the pairwise matrix at varying tolerance ($\Pi_{n}$, $n=5,10,20$). The high mutual information values for the MD datasets reflect the geometrically constrained conformational space that can be explored by the embeddings and suggest potential for shortcut learning.}
    \label{tab:artefact}
\end{table}

Specifically, high correlation is found with sum of pairwise distances, $\Sigma_{+}$, total space writhe $\Omega_{+}$, maximal pairwise distance $\text{M}$, and the number of peaks (connected components) on the pairwise matrix at varying tolerance $n$, $\Pi_{n=10}$ and $\Pi_{n=20}$. 
In contrast, effectively zero correlation is measured in all of the tested geometric features on the \acr{GEOKNOT}-generated dataset. 
We can thus conclude that MD (LAMMPS) sampled data do not cover geometrically complex long-range entanglement, as otherwise the sum of pairwise distances $\Sigma_{+}$ would have remained uninformative toward topological classification. On the contrary, \acr{GEOKNOT}-generated dataset displays very low mutual information irrespectively of the probe, meaning that the space of geometric embeddings of the knots is sampled more uniformly.

\subsection*{Neural Networks can use geometric features to shortcut learn knot topology} 

Having demonstrated that our shortcut probes are correlated with knot topology in the MD-generated dataset, we now want to determine whether ML models can use these correlated features to achieve high accuracies. Specifically, in this section we compare the performance of models trained on different datasets, i.e. MD (LAMMPS) versus \acr{GEOKNOT}, and on different input features, i.e. curve coordinates, writhe and the functionals with high mutual information. The results summarised in Table~\ref{tab:MLaccuracies} and show that the NNs are using the shortcut probes we identified above to infer knot label from the curves' geometry, rather than learning to compute topological invariants. This is possible because MD datasets (both low and high temperature ones) display highly topology-correlated geometric features, and the probes we identified are especially correlated in the low temperature regime. Indeed, training NNs using solely the shortcut probes ($\Sigma_+$, $\Omega_+$, $\dots$) yields 99.9\% classification accuracy in the MD low temperature dataset (see third column in Table~\ref{tab:MLaccuracies}). 

\begin{table}[hbt!]
    \centering
    \begin{tabular}{c|c|c|c|c|c|}
     \textbf{DATA} & \textbf{Coordinate Acc.} & \textbf{Writhe Acc.} & \textbf{Shortcut Probes Acc.} & \hspace {1em} $\bm{\tau}_{\textbf{coord}}$ \hspace {1em} & \hspace {1em} $\bm{\tau}_{\textbf{wr}}$ \hspace {1em} \\ [0.5ex] 
         \hline\hline
        $\text{MD}^{\text{ Low T}}$ &  99.9 & 99.9 & 99.9 & 1.00 & 1.00 \\
        $\text{MD}^{\text{ High T}}$ &  99.9 & 96.7 & 83.1 & 0.83 & 0.86 \\
        \acr{GEOKNOT} & 49.9 & 67.2 & 62.3 & - &  0.93 \\ [0.5ex] 
    \end{tabular}
    \caption{\textbf{ML classification accuracy reveals shortcut learning.} Summary of models and their best accuracies for various input data. The dramatic drop in accuracy for \acr{GEOKNOT} reflects the models inability to classify topology of curves with broad distributions in geometric features. The ``shortcut probes acc.'' column reports accuracy of ML models solely trained on the shortcut probes identified by our algorithm. Notice that $\bm{\tau}_{coord}$ is omitted for \acr{GEOKNOT} data because the coordinate model accuracy is random ($\simeq 50\%$), meaning $\tau$ is not informative about shortcut reliance in this regime.}
    \label{tab:MLaccuracies}
\end{table}

To quantitatively measure the extent of which NNs rely on geometric probes to make correct predictions on knot topology, we measure the ``shortcut learning index'' $\tau = m_a/m$ as the ratio of the accuracies obtained from the ML models when trained on the shortcut probes ($m_{a}$) to the accuracy of ML models trained on the untransformed data ($m$). 

As one can appreciate from Table~\ref{tab:MLaccuracies}, columns 4 and 5, training ML models on geometric features that are flagged by our shortcut probe as ``highly  correlated'' yields similar accuracies to models trained on local writhe matrices and coordinate data, particularly in the low temperature dataset. The ``shortcut learning index'' values, $\tau = m_a/m \simeq 1$, suggest that the models are highly likely to be ``shortcut learning'' topology from these geometric features. In other words, these geometric probes are sufficient for explaining low temperature MD data classification, however, cannot be used to completely explain the high accuracy observed at higher temperatures, which may employ more complicated geometry functionals.

Finally, to identify which features are most important to achieve the high classification accuracy in the MD-trained models, we perform a saliency analysis by back-propagating the class score to the input data in order to identify which features are the ones that most influenced the prediction. 

\begin{figure}[ht!]
    \centering
    \includegraphics[width=1.0\linewidth]{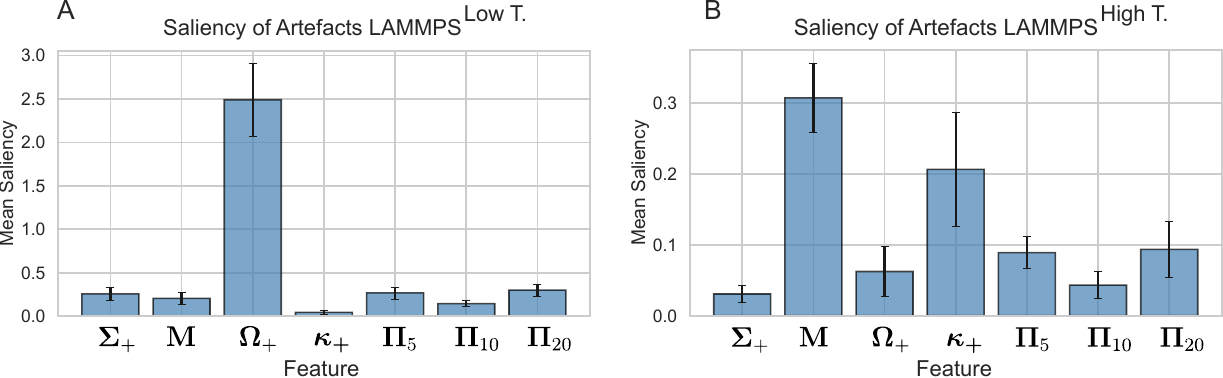}
    \caption{\textbf{Saliency analysis identify key geometric probes for correct classification}. Saliency analysis on ML models trained on geometric probes using both low (A) and high (B) temperature MD (LAMMPS) datasets. At low temperature the $\Omega_+$ feature is by far the most important for correct classification, suggesting that total space writhe is a dominant shortcut feature.}
    \label{fig:figure_3.5}
\end{figure}

We observe that $\Omega_{+}$, i.e. the total space writhe, affects the classification decision substantially more than any other feature in the low temperature dataset (see Figure~\ref{fig:figure_3.5}, left). Again, this is evidence for constrained sampling in the MD (LAMMPS) dataset. Specifically we understand this result as due to the fact that most of the sampled knot conformations in MD are rather minimal perturbations from an \emph{ideal} conformation and that they do not contribute to adding meaningful features in the pairwise distance matrix. 
In other words, fluctuations mostly appear as short length scale twists and very infrequently occur at a meaningful length scale. In fact, the highly separable measures of long-range entanglement distributions in both low and high temperature MD regimes (see Figure~\ref{fig:figure_3}C) further supports this argument.

\subsection*{Topologically invariant deformations reveal reliance of ML models on geometric shortcuts} 

The central question of this paper and of previous works in this area is whether ML models can learn actually understand the topology of a specific embedding rather than correlated geometric features. Interestingly, if the ML model understood topology rather than geometry, then the model would be robust to ambient isotopy, that is, topology preserving deformations. Indeed, the question of whether an ML model can learn topology can be turned into a test of whether the model predictions are invariant under Reidemeister moves. 

To test this invariance we continuously deform the test sample and perform so-called ``ablations'' that modify the embedding whilst keeping the topology fixed. Since our results so far point to the fact that the ML model is shortcut learning through geometric hidden features, we hypothesise that the predictions of the ML models trained on MD-generated data will not be invariant under continuous deformation of the test set. 

For the test set, we generate a continuous path of embeddings $\{X(s)\}_{s\in[0,1]}$ evolved using KnotPlot~\cite{scharein2002interactive} by minimizing the symmetric energy. Specifically, we choose unknot samples from \acr{GEOKNOT} that the models trained on the MD dataset failed to classify (predicted $3_1$ instead). These samples tend to have ``high complexity'', that is, they may have substantially high total space writhe or crossing number in comparison to samples in the MD datasets. 
We sample different embeddings along the evolved path $s\in[0,1]$, track the response of the classifier, $\hat{p}(s)=\mathbb{P}_\theta(0_{1}\mid X(s))$, and finally plot the prediction probability as a function of a chosen geometric functional $G(s)=\phi(X(s))$. 
We specifically focus on total space writhe $\Omega_{+}$ as this is a continuous, interpretable global feature of the embeddings geometry. For topological consistency we ensure that the topology remains unchanged at each time-step by checking the Alexander polynomial of the embedding. 

\begin{figure}[t!]
    \centering
    \includegraphics[width=0.9\linewidth]{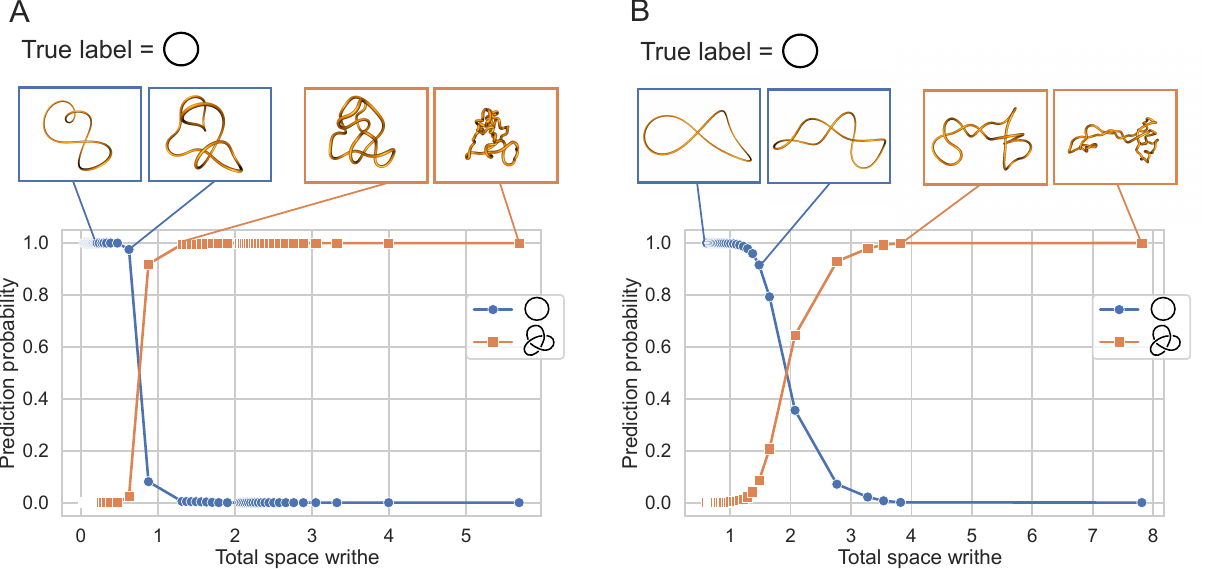}
    \caption{\textbf{ML model prediction is not invariant under ambient isotopy.} In \textbf{A} and \textbf{B} we show two examples of $0_1$ embeddings generated by  \acr{GEOKNOT} that are misclassified as $3_1$ by ML models trained on MD data. The embeddings are minimised by KnotPlot until all non-topological self-crossings are removed. In \textbf{A} the model is trained on the XYZ coordinates of MD data. In \textbf{B} the model is trained on the writhe matrix of MD data. Note, in both plots a smoothed plot of the knot embedding is provided for visual clarity.}
    \label{fig:figure_4}
\end{figure}

In Figure \ref{fig:figure_4}) we report two examples of curves from \acr{GEOKNOT} that are classified as $3_1$ by the models trained on the MD dataset as they have large total space writhe. However, during minimisation these curves lose writhe and self-crossing, eventually minimising to simple unknots. The MD-trained ML models switch prediction when the total space writhe is below 3, consistent with the model utilizing a low-dimensional geometric shortcut to predict the knot topology, rather than learning the topology itself.

\subsection*{Evidence of shortcut learning on published models}

To further generalize the main results of this paper, in this Section we test published models using published architectures and weights from Refs~\cite{sleiman2024geometric, braghetto2025variational, zhang2025recognizing}. 

First, we validate validating our implementation and data by testing our MD-generated datasets (both low and high temperatures) with different models. As one can appreciate in Table~\ref{tab:OtherML} (top two rows), all three published models classify our datasets with very high accuracy. It should also be noted that the representation of input differs between the models above, Sleiman et. al. train models based on the `writhe matrix' representation as input, whereas the Braghetto and Zhang models are trained using coordinate representations, supplemented with `bond vectors' in the latter. We pass the appropriate input in all cases. 

Having benchmarked our implementation of the existing models, we now test their performance on the \acr{GEOKNOT} dataset. As one can appreciate in Table~\ref{tab:OtherML} (bottom row), all existing models do not perform well. 
Specifically; the models from Sleiman et. al.~\cite{sleiman2024geometric} and Braghetto et. al.~\cite{braghetto2025variational} predict the output to be one of 5 knot types ($0_1, \cdots 5_2$), whereas Zhang et. al.~\cite{zhang2025recognizing} predict the knot type to be one of 8 ($0_1, \cdots 6_3$). Interestingly the observed accuracies achieved by the respective models do slightly better than random chance.

Our results therefore suggest that existing published ML model all suffer from shortcut learning, as they fail to predict knots that display geometrically complicated embeddings with large values of writhe and non-local self-crossings.

\begin{table}[bh!]
    \centering
    \begin{tabular}{c|c|c|c|}
     \textbf{MODEL} & \hspace {1em} \textbf{Sleiman et. al.}~\cite{sleiman2024geometric}\hspace {1em} & \hspace {1em} \textbf{Braghetto et. al.}~\cite{braghetto2025variational}\hspace {1em} & \hspace {1em} \textbf{Zhang et. al.}~\cite{zhang2025recognizing}\hspace {1em} \\ [0.5ex] 
     \textbf{REPRESENTATION} & \hspace {1em} \textbf{Writhe matrix}\hspace {1em} & \hspace {1em} \textbf{Coordinates}\hspace {1em} & \hspace {1em} \textbf{Coordinates \& bond vectors}\hspace {1em} \\ [0.5ex] 
         \hline\hline
        $\text{LAMMPS}^{\text{ Low T}}$ &  100.0 & 100.0 & 97.7 \\
        $\text{LAMMPS}^{\text{ High T}}$ &  96.7 & 100.0 & 97.2 \\
        \acr{GEOKNOT} & 52.3 & 31.6 & 14.4 \\ [0.5ex] 
    \end{tabular}
    \caption{\textbf{Testing existing ML models for knot classification.} Classification accuracy of ML models in current literature against our MD (LAMMPS) datasets and \acr{GEOKNOT} dataset.}
    \label{tab:OtherML}
\end{table}

\subsection*{The \acr{GEOKNOT} dataset can be topologically learned through Vassiliev invariants} 

Having shown that existing ML models break down and cannot accurately classify our \acr{GEOKNOT} dataset, we ask whether this dataset can be learned through coordinate data and writhe matrices.
To do this, we now employ the writhe matrix representation to compute Vassiliev (or finite-type) invariants on the \acr{GEOKNOT} data.

Vassiliev invariants can be approximated using only writhe matrix input. Indeed, it is possible to adopt the lens of perturbative 3-dimensional Chern-Simons theory~\cite{bar1995perturbative}, where these invariants can be represented as through ``Feynman diagrams'' according to a set of rules (described in Appendix~\ref{appendix:D}) and eventually expressed in integral formulations. 
Specifically, the integral expression of the second Vassiliev invariant, $\mathcal{I}_{V_{2}}$, is given
{\footnotesize
\begin{align*}
\mathcal{I}_{V_{2}} 
&= \frac{A}{16\pi^{2}}\int d_{x_{1}, ..., x_{4}} 
  \epsilon_{ijk}\epsilon_{lmn}
  \frac{(r(x_{1}) - r(x_{3}))^{k}}{|r(x_{1})-r(x_{3})|^{3}} 
  \frac{(r(x_{2}) - r(x_{4}))^{n}}{|r(x_{2})-r(x_{4})|^{3}} 
  \dot{r}^{i}(x_{1})\dot{r}^{j}(x_{2})\dot{r}^{l}(x_{3})\dot{r}^{m}(x_{4}) \\
& \quad
  + \frac{B}{128\pi^{3}}\int dx_{1, 2, 3}\int_{\mathbb{R}^{3}} d^{3}z \epsilon^{i'j'k'}\epsilon_{ii'i''}\epsilon_{jj'j''}\epsilon_{kk'k''} 
  \frac{(r(x_{1}) - z)^{i''}}{|r(x_{1})-z|^{3}}
  \frac{(r(x_{2}) - z)^{j''}}{|r(x_{2})-z|^{3}}
  \frac{(r(x_{3}) - z)^{k''}}{|r(x_{3})-z|^{3}} 
  \dot{r}^{i}(x_{1})\dot{r}^{j}(x_{2})\dot{r}^{k}(x_{3})
\end{align*}}
where $r(x_{i})$ is the three-dimensional position of segment $x_i$ in the embedding. 
The data constituting the writhe matrix, $\Omega$ (equation~\ref{eq:ab_prop}),  
represents a discretized, segment-averaged approximation to the Gauss linking 2-form such that an algebraic manipulation of the entries of the matrix returns a diagram as follows
\begin{center}
\begin{tikzpicture}
    \draw[thick] (2, 0) circle(0.7cm);
    \draw[dashed, thick] (1.5,0.5) -- (2.5,-0.5);
    \draw[dashed, thick] (2.5,0.5) -- (1.5,-0.5);
    \node at (5, 0) {$=\sum_{i<j<k<l}^{N} \Omega[i][k]*\Omega[j][l]$};
\end{tikzpicture}    
\end{center}
where $\Omega[i][j]$ is the writhe value computed between segments $i$ and $j$.  

We emphasize that this does not constitute a constructive or general proof that writhe matrices determine Vassiliev invariants; rather, it supports the idea that low-order finite-type information is extractable from writhe matrices via explicit diagram-inspired contractions.

\begin{figure}[hbt!]
    \centering
    \includegraphics[width=0.95\linewidth]{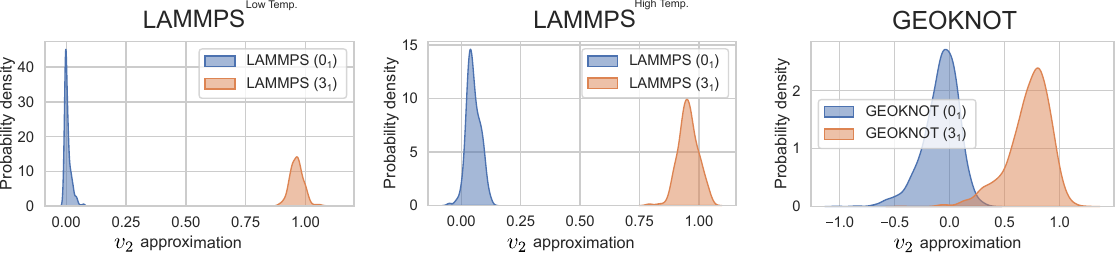}
    \caption{$v_{2}$ approximation from writhe matrix over the low and high temperature LAMMPS and GEOKNOT datasets using writhe matrix data.}
    \label{fig:vassiliev}
\end{figure}

Indeed, running this computation on the \acr{GEOKNOT} dataset returns values that strongly correlate with the second Vassiliev invariant, for $0_1$ and $3_1$ samples as $-0.07\pm0.21$ and $0.72\pm0.25$, respectively, yielding near 100\% ($\approx98.3\%$) classification accuracy (Figure~\ref{fig:vassiliev}).

This demonstrates that the writhe representation encodes nontrivial low-order finite-type information, but that standard ML models fail to recover the corresponding algebraic structure beyond first-order, $\Sigma_{i<j} \Omega[i][j]$.

This available, rule based, algebraic computation lead us to wonder if, sufficiently large ML models could eventually learn to reconstruct Feynman diagrams as above and in turn to recover known diagrammatic structures associated with topological invariants.. Even more exciting would be the prospect of ML models to learn new algebraic manipulations of a given input to yield new topological motifs. 
\section{Discussion and Conclusions}\label{sec:discussion} 

In this paper we have re-assessed the recent successes of ML models in classifying complex knots from different types of embeddings~\cite{vandans2020identifying,sleiman2024geometric,braghetto2023machine, braghetto2025variational, zhang2025recognizing}.  We find evidence that all published and most recent ML models exhibit so-called ``shortcut learning'', whereby the models learn to classify knots not because they understand topology, but rather because they detect geometric features that are highly correlated with knot topology in the given datasets (Figs.~\ref{fig:introduction}-\ref{fig:figure_2}). 

Indeed, we found that the impressive accuracy achieved by ML models in existing knot classification challenges ($> 99\%$) is limited to datasets that are generated using Molecular Dynamics (MD) simulations to sample knot embeddings. We hypothesized that due to the force fields in MD simulations, knot embeddings sample a limited conformational space and therefore display geometric features that are highly correlated to the underlying topology. This was confirmed by measuring the distributions of a selected number of ``geometric functionals'' (see Figs.~\ref{fig:figure_3}-\ref{fig:figure_3.5}), which are clearly separable and therefore strong candidates for shortcut learning.  Inspired by this, we propose to use ``shortcut geometric probes'' in future works in this area to analyse datasets before using them to train ML models.  

To further explore this idea, we created a new biased sampling algorithm, \acr{GEOKNOT}, able to generate knotted configurations that are not restricted in geometric space and thus display broad distributions of writhe and other geometric features that are otherwise strongly restricted in MD simulations due to standard energy potentials. 
We find that this new dataset cannot be learned by current ML models as they fail to return previously seen accuracies. Further, we find that models trained on LAMMPS data fail to generalize to the \acr{GEOKNOT} dataset, further highlighting the prior existence of shortcut learning. Interestingly, minimisation of misclassified unknots from \acr{GEOKNOT} dataset reveals that MD-trained ML models strongly rely on total space writhe to achieve correct classification of knots (Figs.-\ref{fig:figure_3.5}-\ref{fig:figure_4})

We highlight that the findings of this work do not take away the validity of the results of previous investigations. Rather, they specify their scope and domain. 
ML models trained on MD-generated data to identify different topologies serve as valid and accurate probes in the context of physical polymers, proteins or filamentous strings which are subject to particular energetically constrained statistics. 
However, we argue that MD simulations are not the best suited to generate datasets to train ML models to understand topology. Specifically, we argue that ML models trained on MD data are not invariant under continuous deformations of the test samples and therefore cannot encode topological invariants. 

We concluded our work by showing that the writhe matrix potentially encodes sufficient information to learn good approximations of true topological invariants, however the ML models we considered in this work are not powerful enough to learn these invariants. 
We hope that our work will help future investigations in this field and speed up the ML-guided discovery of new topological invariants. 

\section{Data and code availability}
\noindent The code for generating knots of chosen geometries and the \acr{GEOKNOT} dataset is available at: \\
\href{https://github.com/djordjepmihajlovic/GEOKNOT}{https://github.com/djordjepmihajlovic/GEOKNOT}\\

\section{Acknowledgements}
Djordje Mihajlovic’s work was supported by the UKRI Centre for Doctoral Training in Algebra, Geometry and Quantum Fields (AGQ), Grant Number EP/Y035232/1.
Davide Michieletto acknowledges the Royal Society and the European Research Council (grant agreement No 947918, TAP) and the Leverhulme Trust for funding. 
The authors also acknowledge the contribution of the COST Action Eutopia, CA17139. 
We acknowledge insightful conversations with Tudor Dimofte, Renzo Ricca and Alex Klotz.  
We thank Enzo Orlandini, Marco Baiesi and Anna Braghetto for sending us their trained model.
For the purpose of open access, we have applied a Creative Commons Attribution (CCBY) license to any author accepted manuscript version arising from this submission.

\bibliography{libraryv}

\appendix
\section{BFACF and pivot based knot sampling}\label{appendix:A}

To ensure that the geometric state space of the knot embeddings is sufficiently explored, Markov Chain Monte Carlo methods are used to evolve lattice knots in $\mathbb{Z}^{3}$ using the BFACF \cite{de1983new} and pivot \cite{madras1988pivot} algorithms. The BFACF algorithm is a local Monte Carlo algorithm acting on a 1D polygonal chain in $\mathbb{Z}^{3}$ consisting of one of the following update rules acting on a randomly chosen vertex of the polygonal chain;
\begin{enumerate}
    \item\begin{center}
    \begin{tikzpicture}
    \draw[thick] (0,0) -- (1,0);
    \fill (0,0) circle (2pt);
    \fill (1,0) circle (2pt);
    
    \draw[thick] (1,0) -- (2,0);
    \fill (1,0) circle (2pt);
    \fill (2,0) circle (2pt);

    \draw[thick] (2,0) -- (3,0);
    \fill (2,0) circle (2pt);
    \fill (3,0) circle (2pt);
    
    \draw[->, thick] (3.5,0) -- (4.5,0);

    \draw[thick] (5,0) -- (6,0);
    \fill (5,0) circle (2pt);
    \fill (6,0) circle (2pt);
    
    \draw[thick] (6,1) -- (7,1);
    \fill (6,1) circle (2pt);
    \fill (7,1) circle (2pt);
    \draw[thick] (6,0) -- (6,1);
    \draw[thick] (7,0) -- (7,1);
    \fill (6,0) circle (2pt);
    \fill (7,0) circle (2pt);

    \draw[thick] (7,0) -- (8,0);
    \fill (7,0) circle (2pt);
    \fill (8,0) circle (2pt);
    
    \end{tikzpicture}
    \end{center}
\vspace{1em}
 \item\begin{center}
    \begin{tikzpicture}
    \draw[thick] (0,0) -- (1,0);
    \fill (0,0) circle (2pt);
    \fill (1,0) circle (2pt);
    
    \draw[thick] (1,0) -- (2,0);
    \fill (1,0) circle (2pt);
    \fill (2,0) circle (2pt);

    \draw[thick] (2,1) -- (3,1);
    \draw[thick] (2,0) -- (2,1);
    \fill (2,0) circle (2pt);
    \fill (2,1) circle (2pt);
    \fill (3,1) circle (2pt);
    
    \draw[->, thick] (3.5,0) -- (4.5,0);

    \draw[thick] (5,0) -- (6,0);
    \fill (5,0) circle (2pt);
    \fill (6,0) circle (2pt);
    
    \draw[thick] (6,1) -- (7,1);
    \fill (6,1) circle (2pt);
    \fill (7,1) circle (2pt);
    \draw[thick] (6,0) -- (6,1);
    \fill (6,0) circle (2pt);

    \draw[thick] (7,1) -- (8,1);
    \fill (7,1) circle (2pt);
    \fill (8,1) circle (2pt);
    
    \end{tikzpicture}
    \end{center}
\end{enumerate}

Acceptance of an update move requires that after each update there is no intersection of vertices, and that the edges remain connected to at least two vertices, these conditions ensure that topology of the system is conserved between updates. The BFACF algorithm is shown to be ergodic for each knot class \cite{van1992ergodicity}, meaning that after sufficient time, the BFACF algorithm should explore the possible geometric configurations for some lattice knot in some bounded $N\times N \times N$ box. However, it is not known how long is required to reach all configurations; additionally, as the algorithm only acts locally, it takes a long time for the embedded polygon to evolve into a completely decorrelated (w.r.t. radius of gyration or another chosen metric) geometry. To increase the speed of polygon geometric evolution, non-local update moves are implemented using the pivot algorithm. Like the BFACF algorithm, the pivot algorithm is a Monte Carlo algorithm that acts on a 1D polygonal chain in $\mathbb{Z}^{3}$. Update moves are defined by randomly selecting two non-equal edges of the chain, the largest arc connecting the two edges is the pivoted by a randomly chosen degree along an axis defined by the shortest distance between the selected edges. 
\vspace{1em}
\begin{center}
    \begin{tikzpicture}
    \draw[thick] (0,0) -- (2,0.5);
    \draw[thick] (0,0) -- (0,2);
    \draw[thick] (0,2) -- (2,2.5);
    \draw[thick] (2,0.5) -- (2,2.5);
    \draw[dashed, thick][red] (0,0.8) -- (2,1.3);
    \fill (0,0.8) circle (2pt);
    \fill (2,1.3) circle (2pt);
    
    \draw[->, thick] (3,1) -- (4,1);

    \draw[thick] (5,0) -- (7,0.5);
    \draw[thick] (5,0) -- (5,0.8);
    \draw[thick] (7,0.5) -- (7,1.3);
    \fill (5,0.8) circle (2pt);
    \fill (7,1.3) circle (2pt);
    \draw[dashed, thick][red] (5,0.8) -- (7,1.3);

    \draw[<-] (6,1.5) arc (0:90:0.7cm) 
    node[midway, above right] {$90^\circ$};
    
    \draw[thick] (5,0.8) -- (6.5,0.8);
    \draw[thick] (7,1.3) -- (8.5,1.3);
    \draw[thick] (6.5,0.8) -- (8.5,1.3);

    \end{tikzpicture}
    \end{center}

As the update moves are non-local, the pivot algorithm allows for efficient exploration of the geometric space of embeddings. However, the pivot algorithm does not conserve the topology of the system and needs to be balanced with methods to ensure that the updated configuration is of a consistent topology. To do this, we utilize KymoKnot \cite{tubiana2018kymoknot} to determine the Alexander polynomial and the second and third order Vassiliev invariants. 
To optimize speed of exploration, the topology is only checked after a set number of pivot update moves, from which it is rolled back if topology is found to be inconsistent. 

\section{Geometric features in the discretized, polygonal setting}\label{appendix:B}
Here we define several states in the geometric state space which we wish to explore using \acr{GEOKNOT}. Examples of knots sampled across the distribution of each feature are included.
\subsubsection{Pairwise distance}
Pairwise distance is simply the sum over the distances between discretized points along the polygonal knot.
\begin{equation}
    D_{polygon} = \sum^{n}_{x=0}\sum^{n}_{y=0} |r(x) - r(y)|
\end{equation}
Such that in the continuous setting
    \begin{equation}\label{eq:pd}
        D = \int_{x}\int_{y>x}|r(x)-r(y)|
    \end{equation}
\subsubsection{Long-range entanglement}
Long-range entanglement is a score tracking how often `non-local' contributions to measures like the writhe occur. This is done by evaluating the number of polygon vertices that are within a certain (small) distance from one another. Here, $q(x, y)$ is the index number difference between $x$ and $y$. Additionally, for our purposes, $d_{L}$ (distance threshold) and $d_{p}$ (sequence threshold) are set to 5 and 10 respectively.

\begin{equation}
    E_{polygon} = \sum_{x=i}^{p}\sum_{y=i+1}^{p} d(x, y), \text{ where } d(x, y) = \begin{dcases*}  
        \text{1 if } |r(x)-r(y)|< d_{L} \text{ and}\\ \text{min}(q(x, y), p - q(x,y)< d_{p}  = 1\\
        \text{0  otherwise}  \\  
    \end{dcases*} 
\end{equation}
\subsubsection{Total writhe}
For a curve in 3-space the total space writhe is defined:
    \begin{equation}\label{eq:writhe}
        \Omega = \frac{1}{4\pi}\oint_{x}\oint_{y}\frac{r(x)-r(y)\cdot(dr(x)\times dr(y))}{|r(x)-r(y)|^{3}}
    \end{equation}
In the lattice setting to ensure efficient evolution of knot sampling we use the computation developed in  \cite{cimasoni2001computing}; notably, this restricts the computation to 4 projections $T_{K}(A_{i})$.
\begin{equation}
    Wr_{lattice} = \frac{1}{4}\sum_{i}^{4}T_{K}(A_{i})
\end{equation}
Here the indicatrix of the cubic lattice divides a hemisphere into connected components $A_{i}$ of area $\frac{\pi}{2}$, and $T_{K}(A_{i})$ is the tait number defined over all (signed) double points $v$ in a projection onto $A$, 
\begin{equation}
    T_{K}(A) = \sum_{v}s(v), \hspace{1em} s(v) = \pm 1 \text{(sign of crossing).}
\end{equation}
To compute this off lattice, we use methods introduced by Banchoff in \cite{banchoff1976self}.
Specifically, for a discretized, polygonal knot the kernel given in equation \ref{eq:writhe} can be calculated with high precision using solid angle formulae on the vertices of the polygonal chain.
In summary, let $\omega_{ij}$ denote the kernel element between the $i^{\text{th}}$ and $j^{\text{th}}$ component along discretized $x_{1}$ and $x_{2}$ respectively. Define points $x_{1},x_{2}$ as the start and end points of discretized segment $i$ and points $x_{3},x_{4}$ as start and end points of $j$, edges of the polygonal knot. We can then define vectors between points $x_{l}$ and $x_{m}$, denoting them $r_{lm}$. Finally let,
$$n_{1} = \frac{r_{13}\times r_{14}}{|r_{13}\times r_{14}|}, \hspace{1em} 
n_{2} = \frac{r_{14}\times r_{24}}{|r_{14}\times r_{24}|}, \hspace{1em} 
n_{3} = \frac{r_{24}\times r_{23}}{|r_{24}\times r_{23}|}, \hspace{1em} 
n_{4} = \frac{r_{23}\times r_{13}}{|r_{23}\times r_{13}|}$$
$$\omega_{ij}^{*} = \text{arcsin}(n_{1}n_{2}) + \text{arcsin}(n_{2}n_{3}) + \text{arcsin}(n_{3}n_{4}) + \text{arcsin}(n_{4}n_{1})$$

It then follows that
$$\omega_{ij} = \omega_{ij}^{*}\text{sign}((r_{34}\times r_{12})r_{13})$$
$$\frac{1}{4\pi}\sum_{i, j}\omega_{ij} = \Omega$$
\subsubsection{Average crossing number}
The average crossing number $ACN$ is simply taken from an unsigned computation of writhe:
    \begin{equation}\label{eq:writhe}
        ACN = \frac{1}{4\pi}\oint_{x}\oint_{y}\Bigg|\Bigg|\frac{(r(x)-r(y))\cdot(dr(x)\times dr(y))}{|r(x)-r(y)|^{3}}\Bigg|\Bigg|
    \end{equation}
So in the polygonal setting
$$\frac{1}{4\pi}\sum_{i, j}||\omega_{ij}|| = ACN$$

\section{Vassiliev invariant Feynman rules}\label{appendix:D}

In Chern Simons theory, the outer loop of the diagram, $X$, is the Wilson loop, a measurement of the holonomy of the gauge connection around a closed loop, here, we can interpret it as the path a chosen knot follows. Specifically, the Feynman diagrams here are a visualization of the perturbative expansion of the Chern Simons QFT;
\begin{equation}\label{eq:Chern-Simons}
    S_{CS}(A) = \int_{M^{3}} \text{tr} (A \wedge dA + \frac{2}{3} A \wedge A \wedge A)
\end{equation}

Diagram 1 represents a `propagator' in each expansion denoted using a dashed line, whereas diagram 2 represents the incident point of propagator on Wilson loop. Diagram 3 represents 3 coincident propagators within the Feynman diagram; for a detailed exposition on this see~\cite{bar1995perturbative}.
\begin{enumerate}
    \item\begin{center}
        \begin{tikzpicture}
            \node at (-1, 0.35) {$r(x_{1}) $};
            \node at (-1, -0.35) {$i$};
            \draw[dashed, thick] (-0.8,0) -- (1.2,0);
            \node at (1.2, 0.35) {$ r(x_{2})$};
            \node at (1.2, -0.35) {$j$};
            \node at (3.75, 0) {$ = \epsilon_{ikm}\frac{(r(x_{1}) - r(x_{2}))^{m}}{4\pi|r(x_{1}) - r(x_{2})|^{3}}$};
        \end{tikzpicture}
    \end{center}
    \item\begin{center}
        \begin{tikzpicture}
            \node at (-1, -0.7) {$i$};
            \draw (0,0) arc (90:270:0.7cm);
            \draw[dashed, thick] (-0.7,-0.7) -- (1.2,-0.7);
            \node at (2.3, -0.7) {$= \dot{r}^{i}(s_{n})$};
        \end{tikzpicture}
    \end{center}
    \item\begin{center}
    \begin{tikzpicture}
        \node at (-0.1, 0.95) {$i'$};
        \draw[dashed, thick] (0,0) -- (0,0.75);
        \node at (-0.7, -0.7) {$j'$};
        \draw[dashed, thick] (0,0) -- (-0.5,-0.5);
        \node at (0.7, -0.7) {$k'$};
        \draw[dashed, thick] (0,0) -- (0.5,-0.5);
        \node at (2.75, 0) {$ = -\frac{1}{6}\int_{\mathbb{R}^{3}}dz\epsilon^{i'j'k'}$};
    \end{tikzpicture}
\end{center}
\end{enumerate}

\end{document}